\documentclass[a4paper,twoside]{article}

\usepackage{epsfig}
\usepackage{subcaption}
\usepackage{calc}
\usepackage{amssymb}
\usepackage{amstext}
\usepackage{amsmath}
\usepackage{amsthm}
\usepackage{multicol}
\usepackage{pslatex}
\usepackage{apalike}
\usepackage{algorithm2e}
\usepackage{graphicx}
\usepackage{blindtext}
\usepackage[bottom]{footmisc}
\usepackage{SCITEPRESS}     

\begin{document}

\title{Towards Developing Ethical Reasoners: Integrating Probabilistic Reasoning and Decision-Making for Complex AI Systems}

\author{\authorname{Nijesh Upreti, Jessica Ciupa \& Vaishak Belle}
\affiliation{The University of Edinburgh, 10 Crichton Street, Edinburgh EH8 9AB, UK}}


\abstract{A computational ethics framework is essential for AI and autonomous systems operating in complex, real-world environments. Existing approaches often lack the adaptability needed to integrate ethical principles into dynamic and ambiguous contexts, limiting their effectiveness across diverse scenarios. To address these challenges, we outline the necessary ingredients for building a holistic, meta-level framework that combines intermediate representations, probabilistic reasoning, and knowledge representation. The specifications therein emphasize scalability, supporting ethical reasoning at both individual decision-making levels and within the collective dynamics of multi-agent systems. By integrating theoretical principles with contextual factors, it facilitates structured and context-aware decision-making, ensuring alignment with overarching ethical standards. We further explore proposed theorems outlining how ethical reasoners should operate, offering a foundation for practical implementation. These constructs aim to support the development of robust and ethically reliable AI systems capable of navigating the complexities of real-world moral decision-making scenarios.}

\onecolumn \maketitle \normalsize \setcounter{footnote}{0} \vfill

\section{Introduction}
\label{sec:introduction}
Computational Ethics seeks to create full ethical agents—autonomous systems capable of translating moral principles and reasoning into actionable, computable frameworks \cite{Moor2006}. These agents aim to optimize justifiable decisions and achieve ethical competency comparable to or surpassing that of humans \cite{Moor1995,ganascia_modelling_2007,Anderson2006,awad_computational_2022}. Significant progress has been made in fairness-focused learning frameworks \cite{rahman_towards_2024,zhang_fairness_2023,islam_differential_2023} and ad hoc approaches to moral reasoning \cite{kleiman2017learning,krarup_understanding_2022,dennis_formal_2016}, with efforts like Kleiman-Weiner et al.’s common-sense moral model \cite{kleiman2017learning} demonstrating the utility of probabilistic reasoning for ethical judgments. These advances provide valuable insights into specific aspects of ethical reasoning.

However, a unified architecture that integrates sophisticated ethical theories with adaptive learning systems remains an open challenge. Autonomous agents must move beyond task-specific reasoning to navigate complex moral scenarios involving beliefs, intentions, and the propositional attitudes of others \cite{belle_knowledge_2023}. The challenge lies in combining structured reasoning with adaptive learning in the face of uncertainty and variability in ethical decisions. This underscores the urgent need for a robust, theoretically grounded framework capable of addressing these complexities systematically.

Specific ethical frameworks, such as Bentham’s theory of Hedonistic Act Utilitarianism \cite{Bentham2003,Anderson2005} and Ross’s prima facie duties \cite{Ross2002}, have historically informed moral reasoning models. However, these approaches often lack the flexibility needed to navigate ethical ambiguities and uncertainties in real-world contexts. More recent works \cite{Lockhart2000,Gilpin2018ExplainingEA} have sought to adapt ethical principles dynamically to situational ambiguities, yet broader integration is required to address the diversity and evolving nature of decision-making environments.

Ethical decision-making depends not only on abstract principles but also on contextual facts, which are often shaped by subjective interpretations. Existing frameworks, such as Kleiman-Weiner et al.'s probabilistic model, provide useful tools for handling moral judgments. However, a comprehensive architecture that accommodates meta-level specifications is still lacking. To address this, it is essential to define the specifications needed for developing ethical reasoning systems. These specifications should combine robust knowledge representation and reasoning (KRR) techniques, logic, and computational embodiment \cite{Levesque1986,Segun2020}. The goal is to develop adaptive, context-aware systems capable of managing the nuanced complexities of real-world ethical dilemmas.

A critical specification is the use of intermediate representations that integrate contextual factors into ethical decision-making processes. These representations break down complex ethical dilemmas into manageable sub-goals, allowing for a structured and layered approach to decision-making. This structure ensures adaptability to dynamic scenarios while preserving alignment with overarching ethical principles. By incorporating intermediate representations, ethical agents can gain the flexibility and precision required to respond effectively to evolving real-world conditions.

In this paper, we explore the challenges and complexities in developing ethical reasoning systems and identify limitations in existing approaches. We outline the necessary ingredients for building a holistic meta-level framework that integrates reasoning, learning, and uncertainty management, supported by a set of proposed theorems defining foundational properties for ethical reasoning systems. These theorems provide a conceptual foundation for designing agents capable of navigating complex decision-making scenarios. By synthesizing existing knowledge and incorporating causal understanding and probabilistic reasoning, we aim to establish the groundwork for constructing ethical agents equipped to address real-world dilemmas in AI systems.

\section{Intermediate representations for KRR}

Intermediate representations within KRR frameworks can take various forms, each contributing distinctively to modeling complex decision-making environments. These representations may be propositional, capturing straightforward truth-functional statements; they may involve first-order logic for detailed object-property relationships \cite{sanner_symbolic_2010} or include action operators and time sequences to represent dynamic processes and temporal reasoning \cite{beckers2022causal}. Different logics are required depending on the structure of the scenario. Some representations account for multi-agent beliefs \cite{ghaderi_towards_2007}, probabilistic uncertainty \cite{davis_symbolic_2020}, and even the belief-desire-intention (BDI) models standard in multi-agent systems \cite{belle_allegro_2015}. These logics help represent agent beliefs, preferences, and temporal factors for more adaptable decision-making. However, this complexity is also necessary to ensure that every significant aspect of a situation can be accurately captured and reasoned about within the KRR framework.

Consider a self-driving car faced with an ethical decision resembling the classic trolley problem \cite{foot1967problem}. The car in the current state of operation must choose between two routes, Route A and Route B, each presenting different ethical considerations based on potential human casualties, physical damage, and the degree of interference required to change its course. Let's say that the car’s decision-making framework is governed by three prioritized principles: \textbf{Minimize human casualties (X)}, \textbf{Minimize physical damage (Y)}, and \textbf{Minimize external interference (Z)}. 

\begin{figure}[t]
 \centering
  \includegraphics[width=0.48\textwidth,height=7cm]{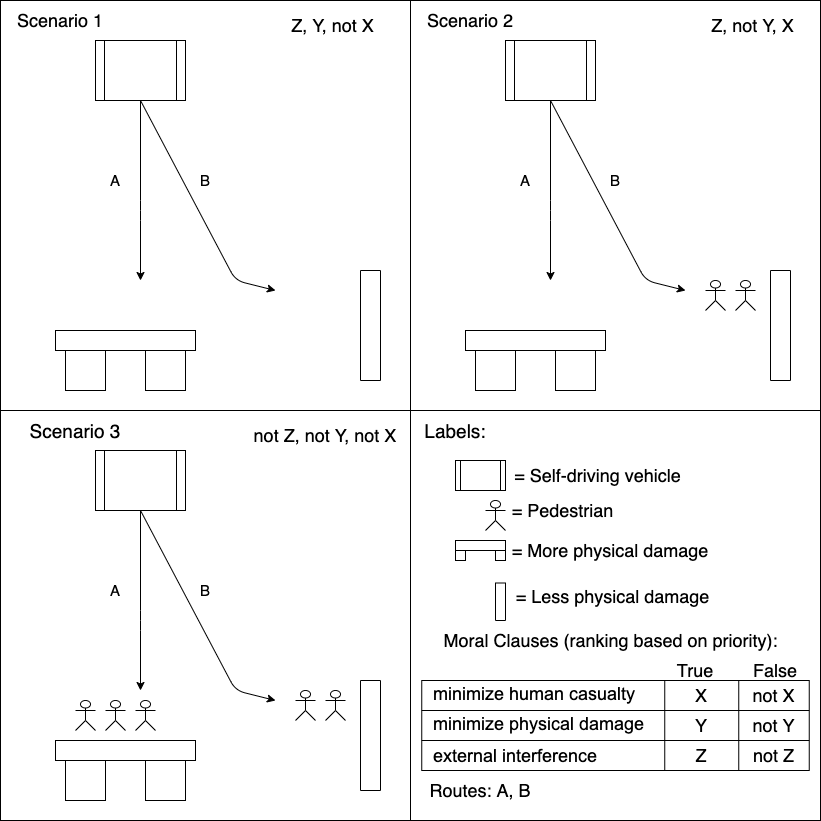}
 \caption{Three scenarios (Scenario 1, Scenario 2, and Scenario 3) each with a choice between two possible routes (A and B) for a self-driving vehicle are depicted in the figure above. }
 \end{figure}

These principles create a structured, yet complex, moral framework for the car’s next move as it calculates the best course of action based on each scenario. The complexity of this ethical calculus is illustrated across four distinct scenarios:

\begin{itemize}
\item \textbf{Scenario 1 (No Pedestrians)}: With no human lives at risk, the vehicle chooses Route B to minimize physical damage (Y) without requiring major interference (Z), as human safety is not a factor.

\item \textbf{Scenario 2 (Two Pedestrian on Route B)}: The presence of a pedestrian on Route B makes minimizing casualties (X) the top priority, leading the vehicle to choose Route A to avoid harm (which also happens to be the default path here).

\item  \textbf{Scenario 3 (Multiple Pedestrians on Both Routes)}: Scenario 3 involves Route A (the default path) with three pedestrians and Route B (requiring intervention) with two pedestrians. The ethical tension here lies between staying on the initial path, potentially sacrificing three lives, and intervening to switch to Route B, saving three lives at the expense of two. This dilemma raises deeper ethical questions: Is it justifiable for the car to intervene in its initial course to save a greater number of lives, even if this requires sacrificing others who are already at risk on Route A?

\item  \textbf{Scenario 4 (Specific Information about Individuals)}: Adding a new layer of complexity, let us consider that the car has specific information about the individuals on each route. Suppose the two individuals on Route A are children, while the three on Route B are adults. Alternatively, perhaps one of the individuals on Route B holds a position of societal importance, such as a medical professional with exceptional skill or highly important public official. The addition of these contextual factors—age, health, and social role—complicates the ethical calculus by introducing nuanced considerations rooted in cultural, ethical, and social norms \cite{andrade2019medical}. 

\end{itemize}

In some ethical frameworks, for example, prioritizing children may be seen as more ethical due to their vulnerability or potential future contributions. In other contexts, societal roles might influence decision-making, with a preference given to individuals who can benefit the broader community. This introduces new questions: Should the vehicle’s ethical framework consider these personal attributes, potentially valuing certain lives differently? Is it ethically justifiable for the system to incorporate societal or cultural values when evaluating lives at risk?

The complexities highlighted by scenarios 3 and 4 reveal the profound challenges in ethical decision-making. Key questions arise, such as: Is it ethical for the vehicle to actively alter its path to save more lives, knowing it risks others in the process? Should characteristics like age or social significance affect the prioritization of lives? These considerations underscore the limitations of a purely casualty-based ethical framework, as individual characteristics and contextual elements often shape ethical perceptions in real-world settings.

Addressing these multifaceted dilemmas requires a structured framework that can balance both contextual factors and ethical objectives. In this context, we introduce ``circumstantial dicta" to represent situational details—such as the vehicle’s initial direction and specific characteristics of individuals—and ``ethical prescripts" to embody moral objectives, such as prioritizing human life or minimizing physical harm. By clearly defining these components, the framework can systematically navigate ethical complexities, adapting its decisions to reflect both individual principles and situational context.

These scenarios illustrate the challenges in creating ethical AI systems capable of handling complex, real-world situations. Ethical decision-making often reflects collective social values and shared understandings of responsibility and vulnerability, not merely individual ethical principles \cite{Ruvinsky2007,kleiman2017learning,belle_knowledge_2023}. For computational ethics, this necessitates frameworks that not only follow programmed rules but also adaptively respond to nuanced human values across varied cultural and ethical landscapes. By integrating detailed contextual information and flexible ethical objectives, AI systems may be better equipped to make informed, ethically sound decisions in diverse and unpredictable environments.

\subsection{Circumstantial Dicta and Ethical Prescripts}
This subsection introduces two key concepts: circumstantial dicta and ethical prescripts. Circumstantial dicta refer to contextual factors influencing an agent's ethical decisions, such as environmental conditions, temporal aspects, societal norms, or other situational details. Circumstantial dicta vary depending on the application, e.g., road conditions in autonomous driving or patient conditions in healthcare. Formally, we define \( C \) as the set of circumstantial dicta, with each \( c_i \in C \) representing a distinct contextual factor. These factors shape the ethical landscape, enabling agents to adapt their decision-making based on situational inputs, ensuring a more comprehensive and pragmatic ethical reasoning framework.

Ethical prescripts encode prioritized objectives such as minimizing harm or promoting fairness. Harm is context-specific, ranging from physical injury to loss of autonomy. Quantifying harm involves utility functions weighted by empirical data or stakeholder priorities, ensuring decisions align with ethical standards across diverse scenarios \cite{krarup_understanding_2022,beckers2023quantifying}. For instance, a prescript in healthcare may prioritize ``do no harm," encouraging low-risk treatment recommendations \cite{goodall2014machine}. In autonomous driving, prescripts might emphasize minimizing harm to pedestrians over vehicle efficiency \cite{bonnefon2016social}. Formally, \( E \) is the set of ethical prescripts, with each \( e_j \in E \) representing a moral directive guiding decisions. Together, circumstantial dicta and ethical prescripts form a framework balancing contextual awareness with ethical principles, enabling consistent, contextually appropriate decision-making.

\subsection{Probabilistic Representation of Ethical and Contextual Factors}

In real-world scenarios, ethical decision-making often involves navigating uncertainties in situational contexts and prioritizing ethical objectives \cite{krarup_understanding_2022,dennis_formal_2016,dennis_practical_2016}. This complexity necessitates a probabilistic approach representing the likelihood and relevance of circumstantial dicta and ethical prescripts \cite{pearl2014probabilistic}. By assigning probabilistic weights to these elements, the framework enables an ethical agent to prioritize situationally appropriate and ethically aligned actions, even amid uncertainty.

\subsubsection{Probability Distribution over Circumstantial Dicta}

The first step in constructing a probabilistic representation is to define a probability distribution over circumstantial dicta. This distribution reflects the likelihood of various contextual factors being relevant to the decision-making process \cite{kleiman2017learning}. For instance, in an autonomous driving context, circumstantial dicta, such as road conditions or pedestrian presence, often interact (e.g., bad weather increasing pedestrian risk). These interdependencies are represented using probabilistic dependency models like Bayesian networks, which identify and resolve feedback loops, ensuring computational feasibility in complex scenarios.

Let  \( C = \{c_1, c_2, \dots, c_n\} \) represent the set of all possible circumstantial dicta relevant to a particular decision context. We define the probability distribution over \( C \) as:
\[
P(C) = \{P(c_i) \mid c_i \in C\}
\]
where \( P(c_i) \) denotes the probability that circumstantial factor \( c_i \) is relevant in a given scenario. Each \( P(c_i) \) quantifies the likelihood of encountering a specific contextual condition, such that the distribution sums to 1:
\[
\sum_{i=1}^n P(c_i) = 1
\]
This distribution enables the agent to interpret the situational elements, assigning higher probabilities to more probable factors. For example, in a healthcare setting, \( c_1 \) might denote ``patient in critical condition" with a specific probability based on historical data, while \( c_2 \) might represent ``patient has a history of heart disease," both contributing to the agent’s contextual awareness and action prioritization \cite{pearl2014probabilistic}.

\subsubsection{Conditional Probability of Ethical Prescripts Given Circumstantial Dicta}

Ethical prescriptions are prioritized in response to circumstantial dicta. The framework uses a conditional probability distribution to assign weights to ethical prescriptions based on their relevance to specific circumstantial factors, allowing for context-sensitive ethical reasoning.

Let  \( E = \{e_1, e_2, \dots, e_m\} \) denote the set of prioritized ethical prescripts. The conditional probability distribution of ethical prescripts given circumstantial dicta is defined as:
\[
P(E|C) = \{P(e_j | c_i) \mid e_j \in E, c_i \in C\}
\]
where \( P(e_j | c_i) \) represents the probability of prioritizing ethical prescript \( e_j \) given the presence of circumstantial factor \( c_i \). This conditional probability captures the degree to which certain ethical priorities should be emphasized in light of particular situational factors.

For instance, in an autonomous driving context, if \( c_i \) represents a high likelihood of pedestrian presence, the ethical prescript \( e_j \) corresponding to "prioritize pedestrian safety" would carry a higher conditional probability, influencing the system toward decisions that emphasize this ethical objective. To ensure a complete probabilistic model, the conditional probabilities over ethical prescripts for each circumstantial factor sum to 1:
\[
\sum_{j=1}^m P(e_j | c_i) = 1 \quad \forall \, c_i \in C
\]
This probabilistic structure ensures that the ethical agent can adjust its moral priorities dynamically based on changing situational indicators, enabling a nuanced and responsive ethical reasoning process \cite{dennis_practical_2016,dennis_formal_2016,pearl2014probabilistic}.

\subsubsection{Joint Probability Distribution of Circumstantial Dicta and Ethical Prescripts}
\label{sec:jointprobdist}
To unify the representation of circumstantial dicta and ethical prescripts within the decision-making framework, a joint probability distribution \( P(C, E) \) is defined. This distribution allows the ethical agent to compute the combined likelihood of encountering specific contextual factors and corresponding ethical prescripts, providing a holistic view of scenario-specific ethical considerations.

The joint probability distribution \( P(C, E) \) is given by:
\[
P(C, E) = P(C) \cdot P(E|C)
\]
where \( P(C) \) represents the independent probability distribution over circumstantial factors, and \( P(E|C) \) denotes the conditional probability distribution over ethical prescripts given circumstantial factors.

This joint distribution forms the basis for calculating expected values associated with various ethical actions, facilitating coherent and contextually relevant decision-making in uncertain environments. For example, in a healthcare setting where \( P(c_1) \) represents a critical patient status and \( P(e_2 | c_1) \) indicates a high priority for life-saving actions, the joint probability \( P(c_1, e_2) \) will strongly influence the agent's decision to pursue life-saving interventions.

The joint probability distribution also supports the calculation of expected utility for each potential action, a process that quantifies the expected value of decisions by factoring in both ethical priorities and contextual conditions \cite{russell2016artificial,von2007theory}. Formally, the expected utility for an action \( a \) can be defined as:
\[
U(a | C, E) = \sum_{i=1}^n \sum_{j=1}^m P(c_i) \cdot P(e_j | c_i) \cdot u(a | c_i, e_j)
\]
where \( u(a | c_i, e_j) \) represents the utility of action \( a \) under circumstantial factor \( c_i \) and ethical prescript \( e_j \). The action that maximizes expected utility is selected as the optimal decision:
\[
a^* = \arg\max_{a \in A} U(a | C, E)
\]
This calculation enables the agent to prioritize actions that align with both ethical imperatives and situational demands, ensuring principled, context-sensitive decision-making.

\subsection{Normalized Collection with Matrix Representation}

A standardized method for structuring ethical priorities and contextual factors supports consistent yet adaptable decision-making across diverse scenarios. Normalized collections, which organize ethical prescripts and circumstantial dicta into reusable matrices, help address shifting contexts and interdependencies, ensuring scalable ethical reasoning \cite{StrehlGhosh2002}.

Each ethical scenario is encapsulated in a matrix \( M = [m_{ij}] \), where each entry \( m_{ij} \) denotes the weighted influence of ethical prescript \( e_i \) in the context of circumstantial factor \( c_j \). Here:
\[
m_{ij} = w(e_i, c_j) \cdot P(e_i | c_j)
\]
where \( w(e_i, c_j) \) represents a baseline weight or relevance of ethical prescript \( e_i \) under the circumstantial factor \( c_j \), and \( P(e_i | c_j) \) is the conditional probability derived from the previous probabilistic model, indicating the likelihood of prioritizing \( e_i \) given \( c_j \).

This matrix structure \( M \) captures a unique ethical profile for each scenario, reflecting the contextualized relevance of ethical prescripts in structured form. Mathematically, this matrix is defined as:
\[
M = 
\begin{bmatrix}
m_{11} & m_{12} & \dots & m_{1n} \\
m_{21} & m_{22} & \dots & m_{2n} \\
\vdots & \vdots & \ddots & \vdots \\
m_{m1} & m_{m2} & \dots & m_{mn} \\
\end{bmatrix}
\]

\subsubsection{Construction and Normalization of the Collection}

The normalized collection, denoted \( \mathcal{N} = \{ M_1, M_2, \dots, M_k \} \), is a standardized set of matrices, each representing an ethical profile drawn from various contexts. Normalization is a crucial process here to allow comparisons and coherence across these profiles, ensuring consistency in the application of ethical prescripts across diverse scenarios.

To achieve normalization, each matrix entry \( m_{ij} \) is scaled to a defined range, such as [0, 1], so that ethical prescripts retain consistent relative significance across contexts. Formally, the normalization process for any matrix \( M_i \in \mathcal{N} \) is expressed as:
\[
M_i^{\text{norm}} = \frac{M_i - \min(M_i)}{\max(M_i) - \min(M_i)}
\]
where \( \max(M_i) \) and \( \min(M_i) \) are the maximum and minimum values in \( M_i \), respectively. This normalized form \( M_i^{\text{norm}} \) ensures that ethical profiles across different contexts can be applied and interpreted uniformly, allowing for adaptability \cite{pearl2014probabilistic,koller2009probabilistic}.

\subsubsection{Ensemble Coding and Clustering of Ethical Profiles for Efficient Contextual Retrieval}

To further enhance adaptability, identifying shared ethical priorities and contextual influences enables the agent to generalize ethical decision-making from one scenario to another. The process includes grouping profiles with similar ethical prescript priorities and contextual factors, allowing the agent to identify and apply ethical principles across scenarios with related patterns. For instance, scenarios in autonomous driving that share similar road conditions, pedestrian presence, and weather may be grouped to ensure ethical priorities are maintained consistently across these conditions \cite{Ariely2001}.

To organize the normalized collection for effective retrieval and application, Cluster Ensemble techniques \cite{StrehlGhosh2002} are employed to arrange matrices by similarity, facilitating efficient access to relevant ethical profiles. This clustering method groups ethical profiles based on shared attributes, simplifying retrieval for contexts that align closely with previously encountered scenarios.

Given clusters \( \{C_1, C_2, \dots, C_k\} \), a new scenario represented by matrix \( M_x \) is assigned to the most relevant cluster \( C_k \) by calculating the distance metric \( d(M_x, M_y) \) between \( M_x \) and each matrix \( M_y \) in the collection:
\[
d(M_x, M_y) = \sum_{i=1}^m \sum_{j=1}^n |m_{ij}^{(x)} - m_{ij}^{(y)}|
\]
where \( m_{ij}^{(x)} \) and \( m_{ij}^{(y)} \) are the entries in matrices \( M_x \) and \( M_y \), respectively. This clustering ensures the ethical reasoner can retrieve and apply ethical principles effectively across varying but related contexts.

By integrating normalization, ensemble coding, and clustering, the normalized collection supports consistent and contextually adaptive ethical reasoning. This organized repository enables the agent to handle a wide variety of real-world scenarios, applying ethical principles in a principled yet flexible manner.

\subsubsection{Non-Deterministic Models for Action Selection}

In ethical reasoning systems, certain scenarios may allow for multiple ethically acceptable actions, each aligned with relevant ethical prescripts but varying in their potential outcomes. To address this complexity, non-deterministic modeling provides a mechanism for probabilistically weighing and selecting actions based on their ethical impact. This model allows the ethical reasoner to account for ethical flexibility while maintaining adherence to prioritized ethical principles.

We define probability distribution over actions \( A = \{a_1, a_2, \dots, a_k\} \) as the set of all potential actions available to the agent in a given scenario. Each action \( a \in A \) is associated with a selection probability \( P(a | C, E) \), which represents the likelihood of choosing \( a \) given a combination of circumstantial factors \( C \) and ethical prescripts \( E \).

The action probability distribution \( P(A | C, E) \) is formally expressed as:
\[
P(A | C, E) = \{P(a | C, E) \mid a \in A\}
\]
where each \( P(a | C, E) \) is calculated by factoring in both the ethical prescripts relevant to the specific context and the probabilities derived from the normalized collection. This distribution ensures that each action’s likelihood is proportionate to its ethical relevance, making ethical decision-making probabilistically flexible across contexts.

\subsubsection{Expected Utility for Action Selection}

To determine the action that best aligns with ethical objectives, an expected utility function \( U(a | C, E) \) is defined for each action \( a \) in the context of circumstantial factors \( C \) and ethical prescripts \( E \) \cite{von2007theory}. The expected utility represents the ethical value or desirability of each action, allowing the reasoner to make decisions that maximize ethical alignment.

The optimal action \( a^* \) is selected based on the expected utility, expressed as (refer to Section \ref{sec:jointprobdist}):
\[
a^* = \arg\max_{a \in A} \sum_{c \in C} \sum_{e \in E} P(c) \cdot P(e | c) \cdot U(a | c, e)
\]
where \( U(a | c, e) \) represents the utility associated with action \( a \) under a specific circumstantial factor \( c \) and ethical prescript \( e \) and \( P(c) \) is the probability of each circumstantial factor, and \( P(e | c) \) is the conditional probability of each ethical prescript given the context.

By maximizing expected utility, this model enables the ethical agent to select an action that aligns with prioritized ethical objectives while factoring in context-driven variability. For example, in an autonomous vehicle scenario, if ethical prescripts emphasize both minimizing harm and respecting pedestrian safety, the expected utility function can weigh these prescripts against circumstantial factors to identify the most ethically appropriate action among available options.

\subsubsection{Multi-Objective Optimization}

In complex ethical decision-making, multiple objectives may need to be optimized simultaneously \cite{keeney1993decisions,coello2006evolutionary}. Multi-objective optimization enables the agent to balance competing ethical prescripts, ensuring that no single ethical directive disproportionately influences the decision at the expense of others.

Formally, this approach introduces a weighted utility function:
\[
U(a | C, E) = \sum_{j=1}^{m} \alpha_j \cdot U_j(a | C, e_j)
\]
where \( U_j(a | C, e_j) \) is the utility associated with action \( a \) in relation to ethical prescript \( e_j \) and \( \alpha_j \) represents the weight assigned to each ethical prescript \( e_j \), reflecting its relative importance in the context of circumstantial factors.

Through this multi-objective framework, the ethical reasoner is equipped to make balanced decisions that satisfy multiple ethical objectives simultaneously. The weight \( \alpha_j \) can be dynamically adjusted based on contextual elements, allowing the system to adaptively balance ethical prescripts as situational demands evolve.

\subsubsection{Integrating Non-Deterministic Models with Normalized Collection}

The non-deterministic model is inherently connected to the normalized collection in that the probabilistic weights assigned to each action reflect the ethical priorities derived from the collection’s matrix representation. Each matrix \( M_i \) in the normalized collection serves as a reference for determining the relative importance of ethical prescripts, and by extension, the probability distribution over actions.

When a scenario is encountered, the corresponding matrix from the normalized collection provides context-specific weights for ethical prescripts, influencing the probability distribution \( P(A | C, E) \) over possible actions. This integration ensures that the ethical reasoner’s decision-making is not only probabilistically flexible but also rooted in consistent ethical principles across diverse contexts.

By combining non-deterministic modeling, expected utility, and multi-objective optimization, this approach enables the ethical agent to navigate complex decision scenarios, balancing ethical principles with situational sensitivity. This layered model serves as the basis for robust, adaptable, and ethically grounded action selection in uncertain environments.

\section{Theoretical Foundations and Desired Theorems}

In this section, we aim to set foundational conditions that provide clarity on how an ethical system behaves under varied circumstances. These principles—consistency, optimality, robustness, convergence, and alignment with human judgment—are essential pillars for ensuring correctness in ethical computations. By defining these properties, we not only articulate the expected behaviors of ethical agents but also create a basis for proving correctness, stability, and effectiveness. We provide sample theorems and strategies for proving their validity, offering a roadmap for substantiating each core attribute of the system’s ethical reasoning with appropriate analytical techniques.\\

\textbf{Theorem 1 (Ethical Consistency)}: Ethical consistency ensures stable prioritization using metrics like cosine similarity or KL divergence, so small changes in context result in proportionate changes in ethical decisions. Specifically, for any two sets of circumstantial dicta, \( C_1 \) and \( C_2 \), where \( C_1 \approx C_2 \), the probability distributions over ethical prescripts should be approximately the same:
\[
\forall C_1, C_2 \in C, \quad \text{if } C_1 \approx C_2, \text{ then } P(E|C_1) \approx P(E|C_2).
\]

This statement ensures that if two situational contexts, \( C_1 \) and \( C_2 \), are nearly identical, then the system’s ethical priorities—represented by the distribution \( P(E|C) \) over ethical prescripts—should not vary disproportionately. Such stability prevents minor contextual differences from causing erratic or unpredictable shifts in ethical decisions, a foundational requirement for reliable ethical decision-making in real-world applications.

To formalize the concept of ``similar” contexts, we introduce a similarity metric \( d(C_1, C_2) \) over the space of circumstantial dicta. This metric quantifies the "distance" between two sets of circumstantial factors, where contexts \( C_1 \) and \( C_2 \) are considered similar if \( d(C_1, C_2) \leq \delta \) for a small threshold \( \delta \). This threshold \( \delta \) represents the maximum allowable contextual variation for two scenarios to be treated as similar in the system’s ethical reasoning process.

To support the claim that \( P(E|C) \) remains stable across similar contexts, a Lipschitz continuity condition \cite{belle_knowledge_2023,rudin1964principles,boyd2004convex} can be applied to \( P(E|C) \). This continuity condition ensures that there exists a constant \( L > 0 \) such that:

\[
\|P(E|C_1) - P(E|C_2)\| \leq L \cdot d(C_1, C_2),
\]

where \( \| \cdot \| \) represents an appropriate norm (such as the \( L_1 \) norm or total variation distance) that measures the difference between the distributions \( P(E|C_1) \) and \( P(E|C_2) \). This condition guarantees that small variations in circumstantial contexts \( C \) produce only proportionally small variations in \( P(E|C) \). The Lipschitz constant \( L \) thus provides a bound on the degree of ethical priority change as the context shifts, establishing a formalized level of stability. The theorem’s significance lies in its ability to confirm that the system’s ethical responses will remain robust under minor contextual changes. By ensuring that similar contexts yield consistent ethical prioritization, we underpin the reliability of the system’s ethical decision-making, thus contributing to predictable, stable, and ethically sound behavior across varying but comparable scenarios.\\

\textbf{Theorem 2 (Decision Optimality):} Optimal decisions maximize expected utility through weighted multi-objective optimization, where utility values are derived from stakeholder-defined priorities or empirical data. This can be mathematically expressed as:
\[
\text{Optimal Decision} = \arg\max_{D} \sum_{i=1}^{n} P(C_i) \cdot U(D, E_i),
\]
where \( U(D, E_i) \) denotes the utility of decision \( D \) under ethical prescript \( E_i \), and \( P(C_i) \) represents the probability of circumstantial dictum \( C_i \).

Here, we want the formulation to capture the principle that the system’s decisions should reflect a rational prioritization of ethical objectives, with each possible decision evaluated based on its alignment with ethical prescripts under specific contextual conditions. By calculating the expected utility of each decision as a weighted sum of the utilities under various circumstantial factors, the system can select the decision that provides the greatest alignment with ethical principles in the given scenario. To frame this mathematically, the expected utility of a decision \( D \), denoted \( \mathbb{E}[U(D)] \), is given by:

\[
\mathbb{E}[U(D)] = \sum_{i=1}^n P(C_i) \cdot U(D, E_i).
\]

Here, the term \( P(C_i) \cdot U(D, E_i) \) represents the contribution of each circumstantial factor \( C_i \) and its associated ethical prescript \( E_i \) to the overall utility of decision \( D \). The optimization problem is thus to identify the decision \( D \) that maximizes this expected utility, selecting the action with the highest ethical alignment.

To ensure an optimal solution for expected utility maximization, certain conditions are required. The utility function \( U(D, E_i) \) must be continuous to avoid abrupt changes in utility due to small variations in decisions or context. Additionally, if the utility function is bounded, it ensures that the optimization process remains well-defined, preventing infinite utility values.

Achieving decision optimality involves several steps. First, the expected utility \( \mathbb{E}[U(D)] \) for each potential decision \( D \) is formulated by calculating a weighted sum of circumstantial factors and ethical prescripts, with each weight reflecting the relevance of specific contextual elements. This expected utility is then framed as an optimization problem to maximize \( \mathbb{E}[U(D)] \), turning the decision selection into a solvable problem. Finally, verifying that conditions such as the continuity and boundedness of the utility function are met ensures that the solution is both practically feasible and theoretically sound, resulting in an optimal, ethically aligned decision. By maximizing expected utility, the system ensures that decisions reflect a balanced consideration of ethical priorities and context.\\ 

\textbf{Theorem 3 (Robustness Under Uncertainty):} An ethical decision-making system is robust if small variations in the probability distributions of circumstantial dicta \( P(C) \) result in proportionately small changes in the decision outcome \( F(D|C) \). This robustness can be formally expressed as:
\[
\forall \epsilon > 0, \exists \delta > 0 \text{ such that if } |P(C) - P(C')| < \delta,
\]
\[
\\ \text{ then } |F(D|C) - F(D|C')| < \epsilon.
\]
where, \( P(C) \) and \( P(C') \) denote the probability distributions over circumstantial dicta for two similar contexts, and \( F(D|C) \) represents the decision function for a given decision \( D \) given the context \( C \).

This expression asserts that for any desired degree of stability \( \epsilon \) in the decision outcome, there exists a threshold \( \delta \) for changes in the probability distribution of circumstantial factors. As long as changes in \( P(C) \) remain within this \( \delta \) threshold, the corresponding decision \( F(D|C) \) will change by no more than \( \epsilon \).

To establish robustness under uncertainty, the approach includes several key steps. First, a sensitivity analysis of the decision function \( F(D|C) \) is conducted by introducing a small perturbation in the probability distribution of circumstantial dicta \( P(C) \) and observing its effect on \( F(D|C) \). This step involves deriving mathematical bounds on the impact of changes in \( P(C) \) on the decision outcome, which can often be achieved through Lipschitz continuity \cite{rudin1964principles,boyd2004convex}. A Lipschitz condition would imply that there exists a constant \( K \) such that:
\[
|F(D|C) - F(D|C')| \leq K |P(C) - P(C')|,
\]
where \( K \) is a bound on the sensitivity of the decision function with respect to changes in \( P(C) \). This condition ensures that the decision function’s response to contextual changes is proportional and does not exceed the threshold of stability. 

In simpler terms, this theorem offers a formal way to assess the system’s ability to remain stable despite small changes in input data. By ensuring this robustness, the system can handle minor variations in context without affecting the reliability or consistency of its ethical decision-making, maintaining trustworthiness across similar but slightly different situations.\\

\textbf{Theorem 4 (Convergence of Ethical Decision-Making):} In a learning-based ethical decision-making system, the system's policy is said to converge if, over time and repeated exposure to similar contexts, the probability of selecting a particular decision given certain circumstantial dicta becomes stable. This convergence is essential for systems that learn from experience, as it ensures that with increasing data, the system’s decisions reach a predictable and consistent pattern, ultimately leading to stable, ethically sound behavior.

The convergence property can be formally stated as:
\[
\lim_{t \to \infty} P(D_{t+1} = D | C_t = C) = 1,
\]
where,  \( D_t \) is the decision made at time \( t \),  \( C_t \) represents the set of circumstantial dicta at time \( t \), and  \( P(D_{t+1} = D | C_t = C) \) is the probability that the system will continue to select decision \( D \) under the circumstances represented by \( C \) as time progresses.

Through the statement, we asserts that, as the system gains more experience with context \( C \) over time, the probability of making a specific decision \( D \) given \( C \) approaches certainty. In practical terms, this means that the system “learns” from its experiences, eventually developing a stable and repeatable pattern of ethical decisions in familiar contexts.

The concept of convergence in decision-making is particularly relevant for systems that adapt based on data, such as those employing reinforcement learning or iterative improvement. Convergence implies that the ethical decision-making policy becomes stable and predictable over time, resulting in a system that behaves reliably and in accordance with predefined ethical standards. For instance, in an autonomous vehicle system, this would mean that as the vehicle encounters similar road scenarios repeatedly, its responses to those scenarios converge, leading to a consistent pattern of decision-making aligned with ethical priorities like pedestrian safety. To demonstrate convergence, the strategy generally involves modeling the learning dynamics as a stochastic process or Markov decision process (MDP), where the probability distribution over decisions given a set of circumstantial factors evolves over time. Proving convergence often requires showing that the policy updates diminish over time, for instance by using a decreasing learning rate in reinforcement learning, ensuring that the updates become smaller as more data accumulates. Additionally, fixed-point theory \cite{hadzic2013fixed} may be applied to identify stable points in the learning process—decisions that do not change as the system iterates through similar scenarios.

Establishing convergence involves several key steps. First, the system’s learning dynamics are modeled as a dynamical system, where decision policy updates are influenced by past outcomes and accumulated experiences. This allows the system’s behavior to be analyzed as it adapts. Next, stability is demonstrated using tools from dynamical systems and probability theory, showing that policy adjustments decrease over time, signaling convergence to a stable state. Finally, limit behavior is proven by showing that, in a given context, the probability of selecting a particular decision stabilizes, ensuring consistent adherence to ethical principles in recurring scenarios.\\

\textbf{Theorem 5 (Alignment with Human Ethical Judgments):} An ethical decision-making model achieves alignment with human ethical judgments when the system's decisions correspond closely to human decisions in similar contexts. Formally, this alignment is measured by the correlation between the model’s decisions, \( D_{\text{model}} \), and human decisions, \( D_{\text{human}} \), expressed as:

\[
\text{Corr}(D_{\text{model}}, D_{\text{human}}) > \theta,
\]

where, \( \text{Corr} \) represents a correlation function that measures the degree of similarity between the decisions made by the model and those made by humans,  \( D_{\text{model}} \) are the decisions made by the model, \( D_{\text{human}} \) are the decisions made by humans, and \( \theta \) is a predefined threshold indicating acceptable alignment.

Here, we suggest that an ethical AI system’s decisions should mirror human ethical reasoning to a significant degree, particularly in complex or morally ambiguous situations. Achieving this alignment ensures that the system’s ethical reasoning aligns with commonly accepted moral standards, enhancing its social acceptability and trustworthiness.

Alignment with human judgments in an ethical AI system is achieved through a blend of empirical data collection and statistical evaluation. First, data on human decisions is gathered across a variety of scenarios and contexts that the model is likely to encounter, allowing for a direct comparison between the model’s choices and human decisions. This comparison involves evaluating the model’s responses in each scenario relative to those made by human subjects, establishing a basis for assessing alignment. Next, statistical methods are employed to calculate the correlation between the model's decisions and human decisions, providing a quantitative measure of alignment. The strength of this correlation serves as an indicator of how closely the model’s ethical reasoning mirrors that of humans. Finally, the correlation is assessed against a predefined threshold \( \theta \), with high correlation values indicating strong alignment. If the correlation falls below this threshold, it suggests that the model may require adjustments to better align with human ethical standards. By aligning its decisions with human judgments, the model gains a measure of validation against human moral reasoning, which is crucial for building ethical AI systems that are consistent with societal expectations and norms.

\section{Discussion}
The central theme of our work so far has been the use of circumstantial dicta, ethical prescripts, and intermediate representations (IRs) to deconstruct complex ethical dilemmas into manageable subgoals, allowing for adaptability in changing contexts while maintaining alignment with core ethical principles. However, practical implementation introduces challenges, particularly in quantifying ethical priorities in real-time, stabilizing decision-making in variable contexts, and addressing diverse real-world scenarios. Ethical dilemmas, such as those embodied in the classic trolley problem, illustrate the difficulty of translating philosophical reasoning into computational processes. These dilemmas are shaped by cultural, societal, and historical constructs, often involving multiple potential equilibria. Computational systems must navigate these moral ambiguities while addressing the trade-offs inherent in real-world decision-making. The challenge lies in ensuring that the underlying ethical reasoning framework remains consistent across various contexts while acknowledging that moral judgments often hinge on subjective, culturally influenced factors.

Advancing ethical reasoning systems necessitates drawing from interdisciplinary insights, particularly from the social sciences. Trust, agent relationships, and affective dimensions play pivotal roles in shaping ethical cognition, especially in multi-agent systems \cite{etzioni_incorporating_2017}. Misalignments in ethical reasoning between agents can erode trust and hinder collective decision-making, a particularly complex issue in scenarios with conflicting ethical goals or shared ethical objectives. Understanding the dynamics of individual agent behavior within collective decision-making processes is essential for building systems that can navigate such complexities. Game theory offers tools for analyzing collective action and ethical decision-making at the group level \cite{ostrom2000collective,coleman2017mathematics}. Concepts like correlated equilibria and coordination games provide valuable frameworks for understanding how individual ethical judgments aggregate into collective moral behavior, especially in multi-agent systems where cooperation or competition is critical.

Incorporating human affective dimensions remains a challenge in computational models but is crucial for aligning AI systems with human values. Insights from neuroeconomics, social sciences, and affective neuroscience provide avenues for integrating emotion-driven, normative aspects into AI systems \cite{Guzak2014}. Embedding these affective elements enhances the alignment of AI systems with human values, fostering adaptability in complex, real-world scenarios. Additionally, meta-ethical considerations, such as the nature of moral principles and their justification, play an essential role in guiding AI design \cite{hagendorff2020ethics}. Should AI systems follow universal ethical principles, or should they adapt to culturally specific moral frameworks? Balancing these approaches ensures that AI systems respect diverse moral values while maintaining consistency in decision-making.

Intermediate representations are pivotal for enabling hierarchical reasoning, which allows ethical decision-making to span multiple levels of abstraction. By decomposing overarching objectives into actionable subgoals, this framework offers a modular approach to managing ethical complexity. However, applying hierarchical reasoning within probabilistic and non-deterministic systems remains an underexplored area that demands rigorous empirical testing to ensure practical applicability. The challenge of aligning system outputs with human ethical judgments persists, especially in probabilistic settings where the uncertainty of outcomes needs to be accounted for. Although probabilistic reasoning formalizes ethical principles, aligning these with human moral standards remains an ongoing challenge. Defining benchmarks, gathering diverse datasets, and setting alignment thresholds are necessary steps for progress. In particular, reinforcement learning, while valuable for optimizing decision-making, carries risks of ethical drift if not carefully monitored. Ensuring that reward functions remain aligned with ethical goals and that updates to policies reflect human ethical intuitions is a critical concern.

Bridging the gap between theoretical frameworks and practical applications requires substantial empirical testing. Validation through simulations, such as autonomous driving scenarios, will benchmark the system’s ethical decision-making against human judgments, helping to identify discrepancies and refine the framework. Adherence to current standards like IEEE P7001 ensures that the system aligns with ethical protocols recognized in the field of autonomous systems \cite{winfield2021ieee}, which contributes to the robustness and practical applicability of the proposed framework. Observing emergent behaviors through simulations, such as those inspired by the Ethics microworld simulator \cite{Kavathatzopoulos2007}, provides critical insights into societal implications. Social simulation frameworks, like those used for studying multi-agent ethical systems \cite{ghorbani2013maia,lopez2012agent,mercuur2019value}, offer valuable methodologies for exploring how individual moral decisions aggregate into collective behavior. These frameworks are instrumental in studying complex multi-agent interactions and the ethical dilemmas that emerge from them.

By integrating interdisciplinary insights, structured reasoning, and probabilistic modeling, this work lays out a foundational framework for the development of adaptive, ethically aligned AI systems. However, realizing this vision will require iterative refinement and extensive empirical validation. Additionally, continued engagement with philosophical and social dimensions is necessary to address the inherent complexities of real-world moral decision-making. The challenges associated with context-dependent values, the quantification of utility, and the management of interrelated ethical prescriptions all point to the need for more comprehensive mechanisms to handle dynamic and uncertain ethical scenarios. Only through such comprehensive efforts can we build AI systems capable of navigating the complexities of moral decision-making in diverse, real-world settings.

\section{Conclusion}
This paper introduces the necessary components for developing computational ethics frameworks, focusing on the integration of probabilistic, non-deterministic, and context-sensitive models. While the approach lays a strong foundation, challenges remain in building adaptive systems capable of navigating the complexities of real-world ethical dilemmas. Moving forward, further refinement, empirical testing, and cross-disciplinary collaboration are needed to ensure these systems can be practically deployed in diverse, unpredictable environments.

\bibliographystyle{apalike}
{\small
\bibliography{example}}

\begin{thebibliography}{}

\bibitem[Anderson et~al., 2005]{Anderson2005}
Anderson, M., Anderson, S., and Armen, C. (2005).
\newblock Towards machine ethics : Implementing two action-based ethical theories.
\newblock In {\em AAAI Fall Symposium}.

\bibitem[Anderson et~al., 2006]{Anderson2006}
Anderson, M., Anderson, S., and Armen, C. (2006).
\newblock An approach to computing ethics.
\newblock {\em {IEEE} Intelligent Systems}, 21(4):56--63.

\bibitem[Andrade, 2019]{andrade2019medical}
Andrade, G. (2019).
\newblock Medical ethics and the trolley problem.
\newblock {\em Journal of Medical Ethics and History of Medicine}, 12.

\bibitem[Ariely, 2001]{Ariely2001}
Ariely, D. (2001).
\newblock Seeing sets: Representation by statistical properties.
\newblock {\em Psychological Science}, 12(2):157--162.

\bibitem[Awad et~al., 2022]{awad_computational_2022}
Awad, E., Levine, S., Anderson, M., Anderson, S.~L., Conitzer, V., Crockett, M., Everett, J.~A., Evgeniou, T., Gopnik, A., Jamison, J.~C., Kim, T.~W., Liao, S.~M., Meyer, M.~N., Mikhail, J., Opoku-Agyemang, K., Borg, J.~S., Schroeder, J., Sinnott-Armstrong, W., Slavkovik, M., and Tenenbaum, J.~B. (2022).
\newblock Computational ethics.
\newblock {\em Trends in Cognitive Sciences}, 26(5):388--405.

\bibitem[Beckers et~al., 2022]{beckers2022causal}
Beckers, S., Chockler, H., and Halpern, J. (2022).
\newblock A causal analysis of harm.
\newblock {\em Advances in Neural Information Processing Systems}, 35:2365--2376.

\bibitem[Beckers et~al., 2023]{beckers2023quantifying}
Beckers, S., Chockler, H., and Halpern, J. (2023).
\newblock Quantifying harm.
\newblock In {\em Proceedings of the 32nd International Joint Conference on Artificial Intelligence (IJCAI 2023)}.

\bibitem[Belle, 2020]{davis_symbolic_2020}
Belle, V. (2020).
\newblock Symbolic {Logic} {Meets} {Machine} {Learning}: {A} {Brief} {Survey} in {Infinite} {Domains}.
\newblock In Davis, J. and Tabia, K., editors, {\em Scalable {Uncertainty} {Management}}, volume 12322, pages 3--16. Springer International Publishing, Cham.
\newblock Series Title: Lecture Notes in Computer Science.

\bibitem[Belle, 2023]{belle_knowledge_2023}
Belle, V. (2023).
\newblock Knowledge representation and acquisition for ethical {AI}: challenges and opportunities.
\newblock {\em Ethics and Information Technology}, 25(1):22.

\bibitem[Belle and Levesque, 2015]{belle_allegro_2015}
Belle, V. and Levesque, H. (2015).
\newblock {ALLEGRO}: {Belief}-{Based} {Programming} in {Stochastic} {Dynamical} {Domains}.
\newblock In {\em Proceedings of the Twenty-Fourth International Joint Conference on Artificial Intelligence (IJCAI 2015)}.

\bibitem[Bentham, 2003]{Bentham2003}
Bentham, J. (2003).
\newblock An introduction to the principles of morals and legislation (chapters i{\textendash}v).
\newblock In {\em Utilitarianism and on Liberty}, pages 17--51. Blackwell Publishing Ltd.

\bibitem[Bonnefon et~al., 2016]{bonnefon2016social}
Bonnefon, J.-F., Shariff, A., and Rahwan, I. (2016).
\newblock The social dilemma of autonomous vehicles.
\newblock {\em Science}, 352(6293):1573--1576.

\bibitem[Boyd and Vandenberghe, 2004]{boyd2004convex}
Boyd, S. and Vandenberghe, L. (2004).
\newblock {\em Convex optimization}.
\newblock Cambridge university press.

\bibitem[Coello, 2006]{coello2006evolutionary}
Coello, C.~C. (2006).
\newblock Evolutionary multi-objective optimization: a historical view of the field.
\newblock {\em IEEE computational intelligence magazine}, 1(1):28--36.

\bibitem[Coleman, 2017]{coleman2017mathematics}
Coleman, J. (2017).
\newblock {\em The mathematics of collective action}.
\newblock Routledge.

\bibitem[Dennis et~al., 2016a]{dennis_formal_2016}
Dennis, L., Fisher, M., Slavkovik, M., and Webster, M. (2016a).
\newblock Formal verification of ethical choices in autonomous systems.
\newblock {\em Robotics and Autonomous Systems}, 77:1--14.

\bibitem[Dennis et~al., 2016b]{dennis_practical_2016}
Dennis, L.~A., Fisher, M., Lincoln, N.~K., Lisitsa, A., and Veres, S.~M. (2016b).
\newblock Practical verification of decision-making in agent-based autonomous systems.
\newblock {\em Automated Software Engineering}, 23(3):305--359.

\bibitem[Etzioni and Etzioni, 2017]{etzioni_incorporating_2017}
Etzioni, A. and Etzioni, O. (2017).
\newblock Incorporating {Ethics} into {Artificial} {Intelligence}.
\newblock {\em The Journal of Ethics}, 21(4):403--418.

\bibitem[Foot, 1967]{foot1967problem}
Foot, P. (1967).
\newblock The problem of abortion and the doctrine of double effect.
\newblock {\em Oxford}, 5:5--15.

\bibitem[Ganascia, 2007]{ganascia_modelling_2007}
Ganascia, J.-G. (2007).
\newblock Modelling ethical rules of lying with {Answer} {Set} {Programming}.
\newblock {\em Ethics and Information Technology}, 9(1):39--47.

\bibitem[Ghaderi et~al., 2007]{ghaderi_towards_2007}
Ghaderi, H., Levesque, H., and Lespérance, Y. (2007).
\newblock Towards a logical theory of coordination and joint ability.
\newblock In {\em Proceedings of the 6th international joint conference on {Autonomous} agents and multiagent systems}, {AAMAS} '07, pages 1--3, New York, NY, USA. Association for Computing Machinery.

\bibitem[Ghorbani et~al., 2013]{ghorbani2013maia}
Ghorbani, A., Dignum, V., Bots, P., and Dijkema, G. (2013).
\newblock Maia: a framework for developing agent-based social simulations.
\newblock {\em JASSS-The Journal of Artificial Societies and Social Simulation}, 16(2):9.

\bibitem[Gilpin et~al., 2018]{Gilpin2018ExplainingEA}
Gilpin, L.~H., Bau, D., Yuan, B.~Z., Bajwa, A., Specter, M.~A., and Kagal, L. (2018).
\newblock Explaining explanations: An overview of interpretability of machine learning.
\newblock In {\em 2018 IEEE 5th International Conference on Data Science and Advanced Analytics (DSAA)}, pages 80--89.

\bibitem[Goodall, 2014]{goodall2014machine}
Goodall, N.~J. (2014).
\newblock Machine ethics and automated vehicles.
\newblock {\em Road vehicle automation}, pages 93--102.

\bibitem[Guzak, 2014]{Guzak2014}
Guzak, J.~R. (2014).
\newblock Affect in ethical decision making: Mood matters.
\newblock {\em Ethics {\&} Behavior}, 25(5):386--399.

\bibitem[Hadzic and Pap, 2013]{hadzic2013fixed}
Hadzic, O. and Pap, E. (2013).
\newblock {\em Fixed point theory in probabilistic metric spaces}, volume 536.
\newblock Springer Science \& Business Media.

\bibitem[Hagendorff, 2020]{hagendorff2020ethics}
Hagendorff, T. (2020).
\newblock The ethics of ai ethics: An evaluation of guidelines.
\newblock {\em Minds and machines}, 30(1):99--120.

\bibitem[Islam et~al., 2023]{islam_differential_2023}
Islam, R., Keya, K.~N., Pan, S., Sarwate, A.~D., and Foulds, J.~R. (2023).
\newblock Differential {Fairness}: {An} {Intersectional} {Framework} for {Fair} {AI}.
\newblock {\em Entropy}, 25(4):660.
\newblock Number: 4 Publisher: Multidisciplinary Digital Publishing Institute.

\bibitem[Kavathatzopoulos et~al., 2007]{Kavathatzopoulos2007}
Kavathatzopoulos, I., Laaksoharju, M., and Rick, C. (2007).
\newblock Simulation and support in ethical decision making.
\newblock {\em Globalisation: Bridging the global nature of Information and Communication Technology and the local nature of human beings}, pages 278--287.

\bibitem[Keeney, 1993]{keeney1993decisions}
Keeney, R.~L. (1993).
\newblock {\em Decisions with multiple objectives: Preferences and value tradeoffs}.
\newblock Cambridge university press.

\bibitem[Kleiman-Weiner et~al., 2017]{kleiman2017learning}
Kleiman-Weiner, M., Saxe, R., and Tenenbaum, J.~B. (2017).
\newblock Learning a commonsense moral theory.
\newblock {\em Cognition}, 167:107--123.

\bibitem[Koller, 2009]{koller2009probabilistic}
Koller, D. (2009).
\newblock {\em Probabilistic Graphical Models: Principles and Techniques}.
\newblock The MIT Press.

\bibitem[Krarup et~al., 2022]{krarup_understanding_2022}
Krarup, B., Lindner, F., Krivic, S., and Long, D. (2022).
\newblock Understanding a robot's guiding ethical principles via automatically generated explanations.
\newblock In {\em 2022 IEEE 18th International Conference on Automation Science and Engineering (CASE)}, pages 627--632, Mexico City. IEEE Xplore.

\bibitem[Levesque, 1986]{Levesque1986}
Levesque, H.~J. (1986).
\newblock Knowledge representation and reasoning.
\newblock {\em Annual Review of Computer Science}, 1(1):255--287.

\bibitem[Lockhart, 2000]{Lockhart2000}
Lockhart, T. (2000).
\newblock {\em Moral uncertainty and its consequences}.
\newblock Oxford University Press, New York.

\bibitem[L{\'o}pez-Paredes et~al., 2012]{lopez2012agent}
L{\'o}pez-Paredes, A., Edmonds, B., and Klugl, F. (2012).
\newblock Agent based simulation of complex social systems.

\bibitem[Mercuur et~al., 2019]{mercuur2019value}
Mercuur, R., Dignum, V., and Jonker, C. (2019).
\newblock The value of values and norms in social simulation.
\newblock {\em Journal of Artificial Societies and Social Simulation}, 22(1):1--9.

\bibitem[Moor, 1995]{Moor1995}
Moor, J.~H. (1995).
\newblock Is ethics computable?
\newblock {\em Metaphilosophy}, 26(1/2):1--21.

\bibitem[Moor, 2006]{Moor2006}
Moor, J.~H. (2006).
\newblock The nature, importance, and difficulty of machine ethics.
\newblock {\em {IEEE} Intelligent Systems}, 21(4):18--21.

\bibitem[Ostrom, 2000]{ostrom2000collective}
Ostrom, E. (2000).
\newblock Collective action and the evolution of social norms.
\newblock {\em Journal of economic perspectives}, 14(3):137--158.

\bibitem[Pearl, 2014]{pearl2014probabilistic}
Pearl, J. (2014).
\newblock {\em Probabilistic reasoning in intelligent systems: networks of plausible inference}.
\newblock Elsevier.

\bibitem[Rahman et~al., 2024]{rahman_towards_2024}
Rahman, M.~M., Pan, S., and Foulds, J.~R. (2024).
\newblock Towards {A} {Unifying} {Human}-{Centered} {AI} {Fairness} {Framework}.
\newblock In {\em Proceedings of the 2024 {International} {Conference} on {Information} {Technology} for {Social} {Good}}, {GoodIT} '24, pages 88--92, New York, NY, USA. Association for Computing Machinery.

\bibitem[Ross, 2002]{Ross2002}
Ross, D. (2002).
\newblock {\em The Right and the Good}.
\newblock Oxford University Press.

\bibitem[Rudin et~al., 1964]{rudin1964principles}
Rudin, W. et~al. (1964).
\newblock {\em Principles of mathematical analysis}, volume~3.
\newblock McGraw-hill New York.

\bibitem[Russell and Norvig, 2016]{russell2016artificial}
Russell, S.~J. and Norvig, P. (2016).
\newblock {\em Artificial intelligence: a modern approach}.
\newblock Pearson.

\bibitem[Ruvinsky, 2007]{Ruvinsky2007}
Ruvinsky, A.~I. (2007).
\newblock Computational ethics.
\newblock In {\em Encyclopedia of Information Ethics and Security}, pages 76--82. {IGI} Global.

\bibitem[Sanner and Kersting, 2010]{sanner_symbolic_2010}
Sanner, S. and Kersting, K. (2010).
\newblock Symbolic {Dynamic} {Programming} for {First}-order {POMDPs}.
\newblock In {\em Proceedings of the AAAI Conference on Artificial Intelligence}, volume~24, pages 1140--1146.
\newblock Number: 1.

\bibitem[Segun, 2020]{Segun2020}
Segun, S.~T. (2020).
\newblock From machine ethics to computational ethics.
\newblock {\em {AI} {\&} {SOCIETY}}.

\bibitem[Strehl and Ghosh, 2002]{StrehlGhosh2002}
Strehl, A. and Ghosh, J. (2002).
\newblock Cluster ensembles {A} knowledge reuse framework for combining partitionings.
\newblock In Dechter, R., Kearns, M.~J., and Sutton, R.~S., editors, {\em Proceedings of the Eighteenth National Conference on Artificial Intelligence and Fourteenth Conference on Innovative Applications of Artificial Intelligence, July 28 - August 1, 2002, Edmonton, Alberta, Canada}, pages 93--99. {AAAI} Press / The {MIT} Press.

\bibitem[Von~Neumann and Morgenstern, 2007]{von2007theory}
Von~Neumann, J. and Morgenstern, O. (2007).
\newblock Theory of games and economic behavior: 60th anniversary commemorative edition.
\newblock In {\em Theory of games and economic behavior}. Princeton university press.

\bibitem[Winfield et~al., 2021]{winfield2021ieee}
Winfield, A.~F., Booth, S., Dennis, L.~A., Egawa, T., Hastie, H., Jacobs, N., Muttram, R.~I., Olszewska, J.~I., Rajabiyazdi, F., Theodorou, A., et~al. (2021).
\newblock Ieee p7001: A proposed standard on transparency.
\newblock {\em Frontiers in Robotics and AI}, 8:665729.

\bibitem[Zhang et~al., 2023]{zhang_fairness_2023}
Zhang, J., Shu, Y., and Yu, H. (2023).
\newblock Fairness in {Design}: {A} {Framework} for {Facilitating} {Ethical} {Artificial} {Intelligence} {Designs}.
\newblock {\em International Journal of Crowd Science}, 7(1):32--39.
\newblock Conference Name: International Journal of Crowd Science.

\end{thebibliography}

\end{document}